\documentclass{article}

    \usepackage[preprint, nonatbib]{neurips_2020}

\usepackage[utf8]{inputenc} 
\usepackage[T1]{fontenc}    
\usepackage{hyperref}       
\usepackage{url}            
\usepackage{booktabs}       
\usepackage{amsfonts}       
\usepackage{nicefrac}       
\usepackage{microtype}      
\usepackage{graphicx}
\usepackage{multirow}
\usepackage{latexsym}
\usepackage{xcolor}

\title{Contrastive Visual-Linguistic Pretraining}

\author{
  Lei Shi\textsuperscript{\rm 1}, Kai Shuang\textsuperscript{\rm 1}, Shijie Geng\textsuperscript{\rm 2}, Peng Su\textsuperscript{\rm 3}, Zhengkai Jiang\textsuperscript{\rm 4} \\
  \textbf{Peng Gao}\textsuperscript{\rm 3}, \textbf{Zuohui Fu}\textsuperscript{\rm 2}, \textbf{Gerard de Melo}\textsuperscript{\rm 5}, \textbf{Sen Su}\textsuperscript{\rm 1}\\
  \textsuperscript{\rm 1}Beijing University of Posts and Telecommunication 
  \textsuperscript{\rm 2}Rutgers University \\
  \textsuperscript{\rm 3}The Chinese University of Hong Kong
  \textsuperscript{\rm 4}Chinese Academy of Sciences 
  \textsuperscript{\rm 5}Hasso Plattner Institute
 }

\begin{document}

\maketitle

\begin{abstract}
Several multi-modality representation learning approaches such as LXMERT and ViLBERT have been proposed recently. Such approaches can achieve superior performance due to the high-level semantic information captured during large-scale multimodal pretraining. However, as ViLBERT and LXMERT adopt visual region regression and classification loss, they often suffer from domain gap and noisy label problems, based on the visual features having been pretrained on the Visual Genome dataset. To overcome these issues, we propose unbiased Contrastive Visual-Linguistic Pretraining (CVLP), which constructs a visual self-supervised loss built upon contrastive learning. We evaluate CVLP on several down-stream tasks, including VQA, GQA and NLVR2 to validate the superiority of contrastive learning on multi-modality representation learning. Our code is available at: \url{https://github.com/ArcherYunDong/CVLP-}. 
\end{abstract}

\section{Introduction}
Language pretraining \cite{devlin2018bert,radford2018improving} has revolutionized Natural Language Understanding (NLU), and strong models such as BERT \cite{devlin2018bert} and RoBERTa \cite{liu2019roberta} are widely used across numerous NLP tasks.
Building on this, Visual-Linguistic Pretraining (VLP) has been proposed to add an extra mask-predict self-supervised strategy for the visual branch \cite{tan2019lxmert,lu2019vilbert}.
Compared with VQA models that are trained from scratch such as DCN \cite{nguyen2018improved}, BAN \cite{kim2018bilinear}, DFAF \cite{gao2019dynamic} and MCAN \cite{yu2019deep},  
VLP relies on a similar network structure as previous methods but can achieve superior performance and better generalization ability thanks to the semantic information acquired from large-scale pretraining.

The two prominent VLP methods LXMERT~\cite{tan2019lxmert} and ViLBERT~\cite{lu2019vilbert} usually perform feature regression or classification for masked visual regions as the pretext task of self-supervised learning. However, we have identified several important problems: 
1) Noisy Labels: $L_2$ feature regression and classification suffer from the noisy annotations in Visual Genome~\cite{krishna2017visual}. 2) Domain Gap: As the visual features are generated by an object detector pretrained on Visual Genome, feature regression and classification of masked regions will make the pretrained visual-linguistic model inherit 
the bias from Visual Genome,
which results in a weak generalization ability on other downstream tasks.
Taking LXMERT as an example, it can generalize better on GQA than on NLVR2 due to the overlap of pretraining and finetuning domains (the visual inputs of GQA \cite{hudson2019gqa} are borrowed from Visual Genome), while the images in NLVR2 \cite{suhr2017corpus} are collected online and consist of entirely different image manifolds compared with the sorts of images used for pretraining.

To solve the aforementioned domain gap and noisy label problems, we propose a novel Contrastive Visual-Linguistic Pretraining (CVLP), which borrows ideas from the popular contrastive learning framework in metric learning to solve the domain bias and noisy label problems. Specifically, CVLP replaces the region regression and classification with contrastive learning, which resolves the above problems. Contrastive learning aims to discriminate between positive examples and negative ones, which does not require any annotation and can solve the noisy label and domain bias problems directly.
However, due to the tremendous memory cost of Transformers \cite{vaswani2017attention}, scaling up the batch size for contrastive learning is difficult. A conspicuous problem of contrastive learning is that the performance is highly constrained by the size of negative examples, which are bounded by the batch size.
Motivated by the idea of memory banks~\cite{wu2018unsupervised,he2019momentum}, we build a dynamic memory queue that caches the contextual features of the previous region and serves as negative examples in contrastive learning. The corresponding cached features drift gradually during training, thus invalidating the previously cached negative features in the memory queue. At the same time, motivated by MoCo~\cite{he2019momentum}, we extract features from the slowly moving query network and store them in the memory queue. When the queue is filled with features, the oldest visual contextual feature will be eliminated from the memory bank. A naive implementation of contrastive learning will fail because the network will learn to discriminate between positive and negative examples quite easily. To solve this problem, we increase feature diversity by adopting a randomly layer-dropping key network \cite{fan2019reducing}.

Our contributions can be summarized as below:
\begin{itemize}
\item We propose a novel contrastive learning framework for visual-linguistic pretraining that solves the domain bias and noisy label problems encountered with previous visual-linguistic pretraining approaches such as LXMERT and ViLBERT.

\item We carry out extensive ablation studies over CVLP to validate our proposed approach. Our CVLP pretraining can achieve significant improvements over a strong baseline (LXMERT), especially when the domain gap between the pretraining and finetuning stages becomes larger. CVLP can surpass the performance of LXMERT on all three datasets (VQA, NLVR2, and GQA).

\end{itemize}

\begin{figure*}[htbp]
 \centering
 \includegraphics[width=0.95\textwidth]{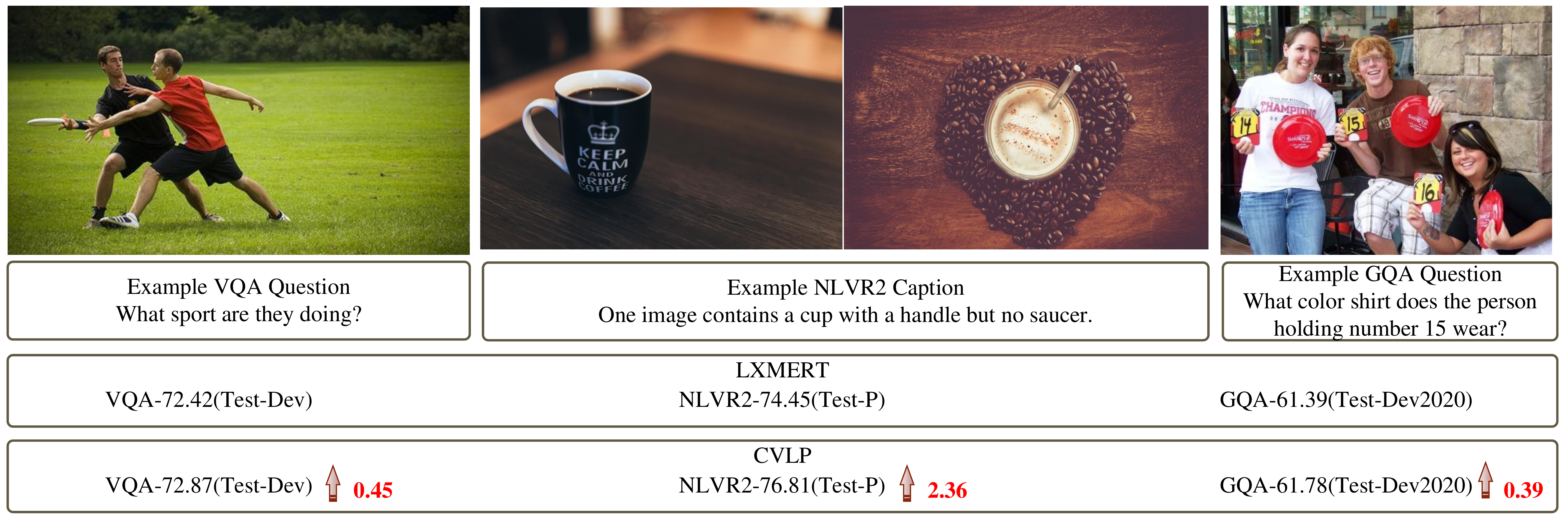}
 \caption{Example question or caption for VQA, NLVR2, GQA datasets. GQA questions are usually longer and more fine-grained than VQA ones, while NLVR2 offers a caption on a pair of images. Our CVLP consistently beats LXMERT across all three vision--language datasets.} 
 \label{all_task}
\end{figure*}

\section{Related Work}
\subsection{Self-supervised Learning in Vision, Language and Multi-modality}
Deep Neural Networks (DNN) trained on ImageNet~\cite{deng2009imagenet} have revolutionized automatic feature representation learning~\cite{krizhevsky2012imagenet}. Compared to supervised training, which incurs a substantial cost for data annotation, self-supervised learning learns useful features automatically by constructing a loss from a pretext task, which does not require human annotation. 
In computer vision, context encoders~\cite{pathak2016context} learn features by image in-painting. Jigsaw~\cite{noroozi2016unsupervised} learns features by predicting the position of permuted features. Kolesnikov et al.~\cite{kolesnikov2019revisiting} carry out a large-scale study of previously proposed self-supervised learning methods and show that the performance of self-supervised tasks varies as the backbone changes. In Natural Language Understanding (NLU), large-scale pretraining with next-word prediction (GPT)~\cite{radford2018improving}, next sentence prediction, or masked word prediction (BERT)~\cite{devlin2018bert}, typically trained with the Transformer architecture~\cite{vaswani2017attention}, has significantly improved the accuracy of NLU, e.g., on the GLUE benchmark~\cite{wang2018glue}. Motivated by the success of self-supervised learning in both vision and language, LXMERT~\cite{tan2019lxmert} and ViLBERT~\cite{lu2019vilbert} have shown that masked words and visual regions can also yield a good visual-linguistic representation.

\subsection{Contrastive Learning}
Contrastive learning is a sub-branch of self-supervised learning,
employing a contrastive loss to learn a representation that is useful in downstream tasks.
The contrastive loss encourages the encoded instance features to be similar to positive keys while keeping away from negative ones.
Different contrastive learning methods adopt different strategies to generate positive and negative keys, which is an essential factor for the quality of learned representation.
\cite{wu2018unsupervised} select the keys from a large memory bank that stores the instance features for the entire training dataset.
\cite{tian2019contrastive,chen2020simple} generate keys using the current mini-batch samples.
MoCo~\cite{he2019momentum,chen2020improved} proposes a momentum encoder to generate the keys on-the-fly and store them in a fixed-size queue.

\subsection{Multi-modality Reasoning}
The backbone of current visual-linguistic pretraining is built upon previous architectures for multi-modal reasoning. Image captioning and VQA~\cite{lin2014microsoft,antol2015vqa,gao2018question,gao2019multi} are two popular tasks that motivate the architecture design for multi-modality fusion. Specifically, attention-based architectures have widely been used in multimodal fusion. Xu et al.~\cite{xu2015show} proposed the first soft and hard attentions, showing that an attention model can yield good performance and interpretability. Yang et al.~\cite{yang2016stacked} proposed a multi-layer attention model by stacking attention models. Besides attention, bilinear models such as MCB~\cite{fukui2016multimodal}, MLB~\cite{kim2016hadamard} and MUTAN~\cite{ben2017mutan} have explored the benefit of channel interactions for multimodal understanding. Subsequently, bottom-up and top-down features~\cite{anderson2018bottom} illustrated the benefit of employing object-level features. Recently, modeling relationships between objects and words as representation learning has been proposed in the DCN~\cite{nguyen2018improved}, BAN~\cite{kim2018bilinear}, DFAF~\cite{gao2019dynamic}, MCAN~\cite{yu2019deep}, QBN~\cite{shi2020multi}, CA-RN~\cite{geng2020character} and STSGR~\cite{geng2020spatio} methods.

\section{Contrastive Visual-Linguistic Pretraining (CVLP)}

\begin{figure*}[htbp]
 \centering
 \includegraphics[width=0.97\textwidth]{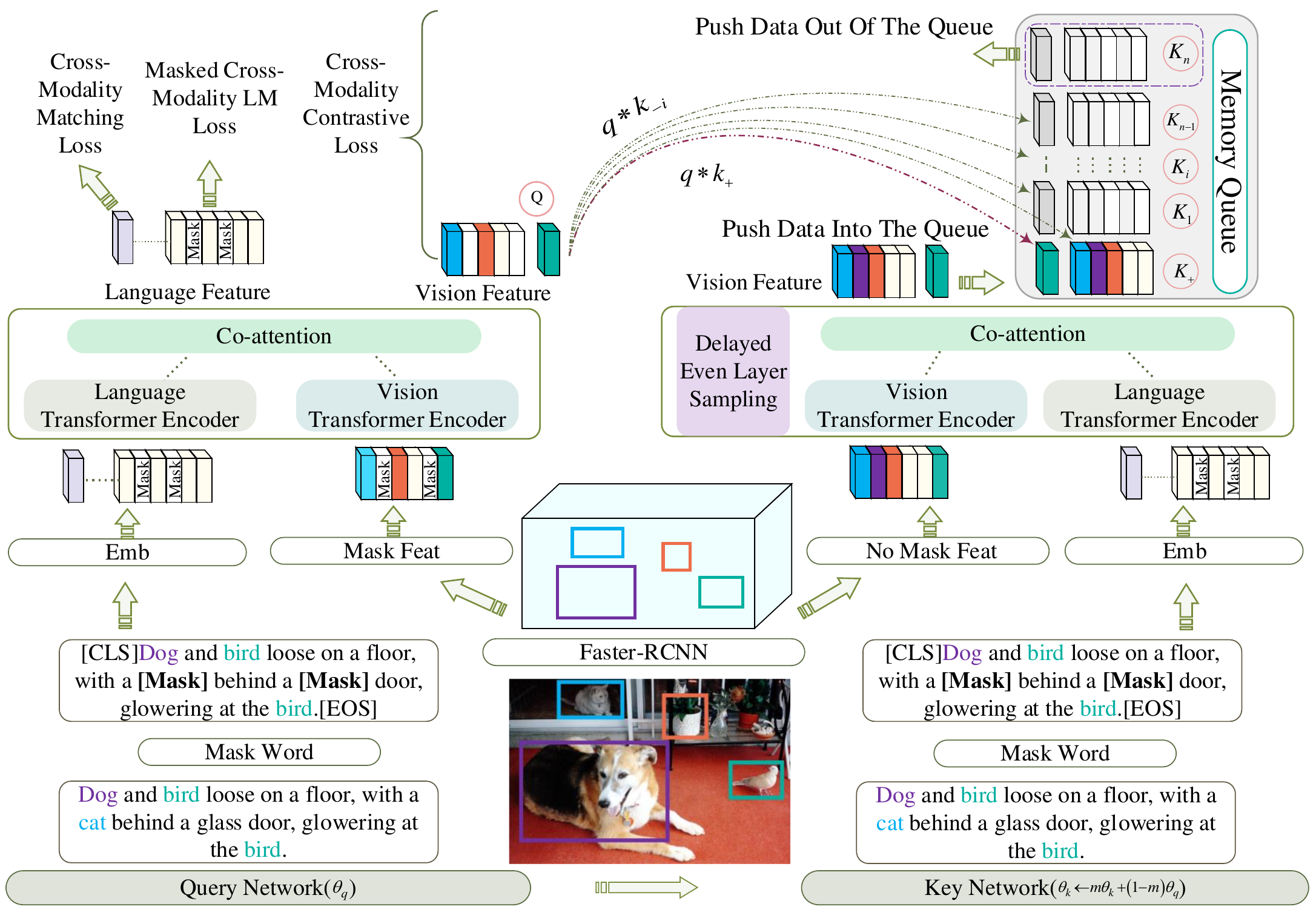}
 \caption{The overall architecture of the proposed CVLP approach. CVLP includes a Query Network, a Key Network and maintains a dynamic memory queue. The entire model is trained with a combination of three cross-modality losses.}
 \label{global2}
\end{figure*}

As illustrated in Figure~\ref{global2}, the architecture of CVLP consists of a Query Network (QueryNet) and a Key Network (KeyNet). They both contain a Language Transformer Encoder, a Vision Transformer Encoder and a Multi-modality Fusion Transformer Encoder. At initialization, KeyNet is copied from QueryNet with the same layers and parameters. The QueryNet produces cross-modality embeddings with a masking strategy applied on both visual and textual inputs, while the KeyNet generates contextualized visual features with masking only applied to textual inputs. The output features of KeyNet are pushed into a dynamic memory queue, which continuously generates negative samples for calculating the Cross-Modality Contrastive Loss. The full CVLP model is trained with a combination of Cross-Modality Masked Language Modeling Loss, Matching Loss and Contrastive Loss. The following subsections are organized as follows: Section~\ref{sec:fusion} introduces how visual and textual features are extracted and fused through self-attention and co-attention strategies, Sections~\ref{sec:lang} and \ref{sec:vis} describe the design of the mask loss for the language branch and the contrastive loss for the visual branch, respectively. Section~\ref{sec:queue_drop} provides further details about the dynamic memory queue mechanism and the droplayer strategy.

\subsection{Multi-modality Fusion}
\label{sec:fusion}
Given image--sentence pairs from a vision--language dataset, we first tokenize each sentence using the WordPieces technique~\cite{wu2016google} and map a token $W_j$ to its corresponding embedding $h_{\rm{emb}}(W_j) \in {\mathbb{R}^{d_w}}$, where $d_w=768$. In addition, visual regions $B \in {\mathbb{R}^{N \times 4}}$ and their features $F \in {\mathbb{R}^{N \times d_o}}$ are extracted by a Faster-RNN~\cite{ren2015faster} detector pretrained on Visual Genome~\cite{krishna2017visual} for each image $I$: $B, F = \mathrm{RCNN}(I)$, where we detect $N = 36$ regions in each image and each region is represented using a feature dimensionality of $d_o=2048$. Then we can calculate the visual inputs $v_i$ and textual inputs $w_j$ of CVLP as follows:
\begin{equation}
\label{init-v-feat-init-w-feat}
     {v_i} = \frac{{{g_{\rm{F}}}\left( {{F_i}} \right) + {g_{\rm{P - ROI}}}\left( {{B_i}} \right)}}{2},
\hspace{1em}
     {w_j} = {h_{\rm{emb}}\left( {{W_j}} \right) + {h_{\rm{P - word}}}\left(P_j \right)},
\end{equation}
where $g_{\rm{F}}$ and $g_{\rm{P - ROI}}$ are two fully-connected layers that map $F_i$ and $B_i$, respectively, to the feature dimensionality $d_w$, while $h_{\rm{P-word}}$ is a positional encoding function for the position $P_j$ of token $W_j$. 

Taking $v_i$ and $w_j$ as inputs, CVLP adopts masking for both QueryNet and KeyNet. For QueryNet, we uniformly choose 15\% of the input textual tokens for replacement. Some of the chosen tokens are replaced by the special \textit{[MASK]} token, while the other tokens are substituted by a random token. For visual regions, we use a different masking strategy: the features of the chosen regions can either be set to zero or be replaced by region features from other images. Different from QueryNet, KeyNet only employs masking on the textual inputs, while keeping all visual region features unchanged. 
KeyNet and QueryNet are initialized to have the same layers and parameters. They both contain 9 layers of Language Transformer Encoder, 5 layers of Vision Transformer Encoder and 5 layers of Multi-Modality Fusion Transformer Encoder. For example, all layers in a KeyNet can be represented as:
\begin{equation}
     \rm{KeyNet} = \left\{ \begin{array}{l}
{r_1},{r_2},{r_3},{r_4},{r_5},\\
{l_1},{l_2},{l_3},{l_4},{l_5},{l_6},{l_7},{l_8},{l_9},\\
c{o_1},c{o_2},co{}_3,c{o_4},c{o_5}
\end{array} \right\},
\end{equation}
where $r_i$ stands for a self-attention layer in the visual branch, $l_i$ stands for a self-attention layer in the language branch, $co_i$ stands for a co-attention layer in the multimodality fusion branch.

The three encoders are implemented by 3 modules, namely, the visual self-attention, language self-attention and visual-linguistic co-attention modules. Visual self-attention performs information fusion between region features by using such features as both key, query and value in the attention model. We denote the key, query and value features for visual as $K_v$, $Q_v$, $V_v$, for language as $K_w$, $Q_w$, $V_w$, respectively. Then the intra-modality information fusion for visual and language features can be denoted as:
\begin{equation}\label{eq:array}
     \widehat v = \rm{Intra}_{v \leftarrow v}\left( {{Q_v},{K_v},{V_v}} \right),
\hspace{1em}
     \widehat w{\rm{ }} = \rm{Intra}_{_{w \leftarrow w}}\left( {{Q_w},{K_w},{V_w}} \right)
\end{equation}
where the attention module of Transformer layer can be denoted as below:
\begin{equation}
    \rm{Attention}(Q,K,V) = \mathrm{Softmax} (Q{K^T}/\sqrt d )V
\end{equation}
After deploying intra-modality information flow for language and visual signals, we invoke an inter-modality fusion module to fuse the information from both language and visual features. The inter-modality fusion process is bi-directional, which includes information fusion from language to vision and vice versa:
\begin{equation}\label{coattention-w-v}
     \widetilde v = \rm{Inter}_{v \leftarrow w}\left( {{Q_v},{K_w},{V_w}} \right),
\hspace{1em}
     \widetilde w = \rm{Inter}_{_{w \leftarrow v}}\left( {{Q_w},{K_v},{V_v}} \right)
\end{equation}
After intra-inter modality feature fusion, we can acquire a multi-modality contextual feature embedding for each word and visual region. A contextual feature encodes the multi-modality interaction in a compact feature vector. The contextual features are used by CVLP for the mask loss in the language branch and the contrastive loss in the visual branch.

\subsection{Mask Loss for Language Branch}
\label{sec:lang}
In the pretraining stage, CVLP performs different pretext tasks compared with LXMERT. CVLP does not contain a supervised learning task and thus is independent of human-annotated labels. For the language branch, we keep masked language modeling and image--sentence matching prediction as two pretext tasks. Mask loss was first proposed by BERT. Subsequent visual-linguistic BERT approaches add a visual feature mask loss besides the masked language modeling loss. This loss masks out the contextual representation obtained in Section~\ref{sec:fusion} and predicts the masked feature using its contextual information. By optimizing the mask loss, the Transformer implicitly learns to encode contextual information, which facilitates the generalization on downstream tasks. 
In CVLP, we only utilize mask loss for the text inputs.
Additionally, we also add a matching loss, which involves a binary Yes/No classification to predict whether the sentence matches the visual feature or not. The mask loss can be formalized as follows:
\begin{equation}
     {\mathcal{L}_{\rm{MLM}}} =  - {E_{w\sim D}}\log {P_\theta }\left( {{w_m}|\widetilde {{w_{/m}}}} \right),
\end{equation}
where $\theta$ denotes the parameters of the Language Transformer Encoder, $w_m$ and $\widetilde {{w_{/m}}}$ are the masked token to be predicted and the contextual tokens that are not masked. The matching loss is defined as:
\begin{equation}
     {\mathcal{L}_{\rm{MATCH}}} =  - {E_{{w_{\rm{CLS}}}\sim D}}\left[ {y\log {P_\theta }\left( {\widetilde {{w_{\rm{CLS}}}}} \right) + \left( {1 - y} \right)\log \left[ {1 - {P_\theta }\left( {\widetilde {{w_{\rm{CLS}}}}} \right)} \right]} \right],
\end{equation}
which is clearly a binary classification task. In the above equation, $w_{\rm{CLS}}$ stands for the \textit{[CLS]} token which encodes the visual-linguistic information for tackling the image--sentence matching pretext task.

\subsection{Contrastive Loss for Visual Branch}
\label{sec:vis}
Contrastive learning performs self-supervised representation learning by discriminating visually similar feature pairs from a group of negative features. Given visual region features extracted by Faster-RCNN, we can obtain a positive query-key pair by feeding such features into both  QueryNet and KeyNet. All region features from other batches are utilized as negative keys. Then we conduct contrastive learning by updating network weights to minimize the following loss:
\begin{equation}
     \mathcal{L}_{\rm{CONTRAST}} =  - \log \frac{\exp \left( {{s^ + }/\tau } \right)}{\exp \left( {{s^ + }/\tau } \right)  + \sum\nolimits_{j = 0}^K {\exp \left( {s_j^ - /\tau } \right)}}
\end{equation}
\begin{equation}\label{posi-negati-sample}
     {s^ + } = \widetilde {v_i^{query}} \cdot \widetilde {v_i^{key + }},
\hspace{1em}
     {s^ - } = \widetilde {v_i^{query}} \cdot \widetilde {v_j^{\rm{memory}\_\rm{queue}}}
\end{equation}
where $\tau$ is the temperature of Softmax, $\widetilde {v_i^{key + }}$ are all positive keys of $\widetilde {v_i^{query}}$, and $\widetilde {v_j^{\rm{memory}\_\rm{queue}}}$ serves as negative examples for calculating ${\mathcal{L}_{\rm{CONTRAST}}}$.
Traditional contrastive learning is constrained by the size of negative examples. In practice, it is time-consuming to acquire a large-sized pool of negative samples. Motivated by Momentum Contrastive (MoCo)~\cite{he2019momentum}, we build a dynamic visual memory queue to store the features generated by the KeyNet. The visual memory queue is empty at first, and features generated by the KeyNet are gradually placed into the queue. 
As training goes on, we can obtain a large visual queue to serve as negative examples. The performance of contrastive learning depends significantly on the feature diversity 
of the visual queue. Once the queue is full, we eliminate the oldest features.
We denote the visual memory queue as:
\begin{equation}
     \rm{memory\_queue} = [\widetilde {v_{b,n}^i}],
\end{equation}
where $\widetilde {v_{b,n}^i}$ represents the visual feature that comes from the $n$-th region of the $b$-th image in the $i$-th iteration batch.
One drawback of visual memory queue is feature drift
during training. As the neural network is updated rapidly, the extracted features may become outdated fairly soon, which invalidates the negative examples stored in the visual queue. To resolve this, we define the weight of KeyNet as a moving average of QueryNet when QueryNet is trained through stochastic gradient descent. The update of the network is denoted as:
\begin{equation}
     {\theta _k} \leftarrow m{\theta _k} + \left( {1 - m} \right){\theta _q},
\end{equation}
where $m$ stands for a momentum value, $\theta_k$ and $\theta_q$ are the parameters of KeyNet and QueryNet respectively.
This form of contrastive learning can achieve superior performance due to the large visual memory queue and the small feature drift during the training progress.

\subsection{Randomly Layer-Dropping Key Network}
\label{sec:queue_drop}
One important factor in training unsupervised representation learning by contrastive learning is to diversify the negative examples. Contrastive learning is highly susceptible to overfitting, thus invalidating the representation learning process. We observe that the contrastive learning loss becomes very small as the training process proceeds, suggesting that overfitting has occurred. We thus increase the diversity of features stored in the visual memory queue through the Randomly Layer-dropping Key Network. The droplayer strategy consists of a random dropout of self-attention and co-attention layers in KeyNet, which can increase the feature diversity and prevent overfitting during the training process of contrastive learning. The randomly layer-dropping Key Network can be defined as follows:
\begin{equation}
     \rm{KeyNet} = \left\{ \begin{array}{l}
{r_1},\rm{SPL}\left( {{r_2}} \right),{r_3},\rm{SPL}\left( {{r_4}} \right),{r_5},\\
{l_1},\rm{SPL}\left( {{l_2}} \right),{l_3},\rm{SPL}\left( {{l_4}} \right),{l_5},\rm{SPL}\left( {{l_6}} \right),{l_7},\rm{SPL}\left( {{l_8}} \right),{l_9},\\
c{o_1},\rm{SPL}\left( {co{}_2} \right),c{o_3},\rm{SPL}\left( {co{}_4} \right),c{o_5},
\end{array} \right\}
\end{equation}
where $\rm{SPL}$ stands for random dropout of a layer or not. As the above equation shows, even layers in the KeyNet may be dropped during pretraining with a sampling probability of 0.5.

\section{Experiments}
In this section, we first introduce the implementation details of the proposed contrastive visual-linguistic pretraining network. Then we conduct extensive ablation studies to demonstrate the effectiveness of the proposed method. CVLP is pretrained on the same dataset as LXMERT, namely MSCOCO and Visual Genome. To assess the learned visual-linguistic features, we conduct finetuning experiments and compare CVLP with state-of-the-art methods on three downstream tasks, i.e., VQA v2~\cite{balanced_vqa_v2}, GQA~\cite{hudson2019gqa} and NLVR2~\cite{suhr2018corpus}.

\begin{figure*}[htbp]
 \centering
 \includegraphics[width=0.92\textwidth]{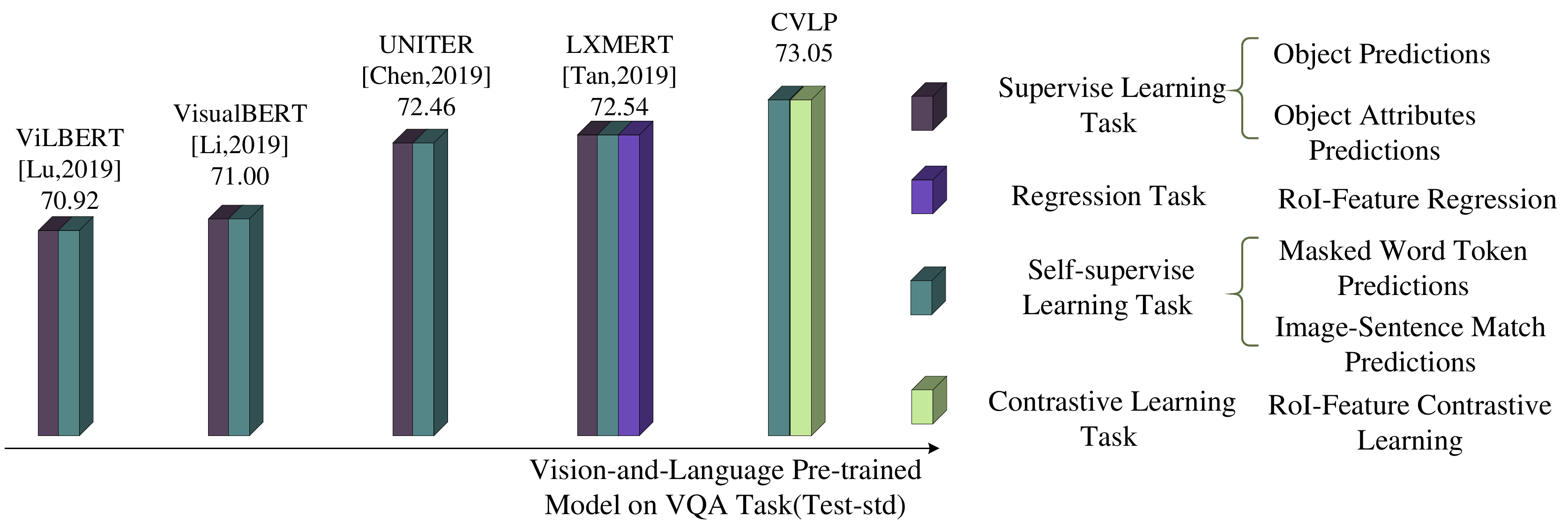}
 \caption{We show the compositions of pretext tasks used by various visual-linguistic pre-training models. Different pretext tasks require different levels of annotations and have multiple effects on downstream tasks. The height of the bars reflects the performance of each method on VQA Test-std.
 }
\label{NCE}
\end{figure*}

\paragraph{Implementation Details.}
Following LXMERT, we pretrain CVLP on the same image--text pairs from MSCOCO~\cite{lin2014microsoft} and Visual Genome~\cite{krishna2017visual}. In the pre-training stage, the batch size is set to 256 and the initial learning rate is 1e-4. During finetuning, the batch size is reduced to 32 and the initial learning rates of downstream tasks VQA, GQA and NLVR2 are 5e-5, 1e-5 and 5e-5, respectively. The temperature $\tau$ in the contrastive loss is set to 0.07. In both pre-training and fine-tuning stages, CVLP is optimized with Adam~\cite{kingma2014adam} on four Nvidia Tesla P100 GPUs.

\subsection{Comparison with State-of-The-Art VLP Methods}
We compare our proposed CVLP with previous visual-linguistic pretraining models, including ViLBERT~\cite{lu2019vilbert}, VisualBERT~\cite{li2019visualbert}, UNITER~\cite{chen2019uniter} and LXMERT~\cite{tan2019lxmert}. The pretraining loss utilized in each specific method is presented in Figure ~\ref{NCE}. All previous methods adopt masked visual region classification and regression. CVLP, in contrast, only needs mask loss on the text modality and contrastive learning loss on visual modality. With the help of this contrastive learning, CVLP achieves better performance on all three downstream tasks compared to previous approaches. In Table~\ref{tab:main}, we can also see that CVLP improves by 2.36\% over the runner-up model UNITER on NLVR2. This improvement is the biggest among CVLP's improvements on all the three datasets, suggesting that CVLP possesses good generalization ability for large domain gap settings.

\begin{table}[!htbp]
\centering
\setlength{\abovecaptionskip}{6pt}
\setlength{\belowcaptionskip}{0pt}
\scalebox{0.92}{
\begin{tabular}{cccc}
\toprule
\multirow{3}{*}{\textbf{Method}}                                             & \multirow{2}{*}{\textbf{VQA}}                                          & \multirow{2}{*}{\textbf{GQA}} & \multirow{2}{*}{\textbf{NLVR2}} \\
                                                                    &                                                               &                      &                        \\ 
                                                                    & \begin{tabular}[c]{@{}c@{}}Test-dev /  Test-std\end{tabular} & Test-dev-2020       & Test-P                 \\ \midrule

Human                                                               & -                                                             & 89.30                & 96.30                  \\ 
Image Only                                                          & -                                                             & 17.80                & 51.90                  \\ 
Language Only                                                       & 44.30 / -                                                         & 41.10                & 51.10                  \\ \hline
LXMERT~\cite{tan2019lxmert}                                                             & \underline{72.42} / \underline{72.54}                                                  & \underline{61.39}                & 74.45                  \\ 
ViLBERT~\cite{lu2019vilbert}                                                            & 70.55 / 70.92                                                  & -                    & -                      \\ 
\begin{tabular}[c]{@{}c@{}}VisualBERT~\cite{li2019visualbert}\\ (w/o Ensemble)\end{tabular} & 70.08 / 71.00                                                 & -                    & 67.00                  \\ 
UNITER~\cite{chen2019uniter}                                                              & 72.27 / 72.46                                                 & -                    & \underline{75.58}                  \\ \hline
\begin{tabular}[c]{@{}c@{}}CVLP\\ (finetune w/o momentum)\end{tabular}            & \textbf{72.77} / \textbf{72.90}                                         & \textbf{61.55}                & \textbf{76.20}         \\ 
\begin{tabular}[c]{@{}c@{}}CVLP\\ (finetune with momentum)\end{tabular}           & \textbf{72.87} / \textbf{73.05}                                         & \textbf{61.78}                & \textbf{76.81}                      \\ \bottomrule
\end{tabular}
}
\caption{Performance comparison between CVLP and state-of-the-art visual-linguistic pretraining approaches on test splits of VQA v2, GQA 2020, and NLVR2. For all datasets, the best accuracy is in bold while the second best accuracy is underlined.}
\label{tab:main}
\end{table}

\begin{table}[!htbp]
\centering
\setlength{\abovecaptionskip}{6pt}
\setlength{\belowcaptionskip}{0pt}
\scalebox{0.92}{
\begin{tabular}{c|cccc}
\toprule
\multirow{2}{*}{\begin{tabular}[c]{@{}c@{}}Momentum value for pre-training\end{tabular}} & \multirow{2}{*}{$m$ = 0.999} & \multirow{2}{*}{$m$ = 0.9999} & \multirow{2}{*}{$m$ = 0.99995} & \multirow{2}{*}{$m$ = 0.99999} \\
                                                                                              &                            &                             &                              &                              \\ \hline
Acc\%                                                                                           & 70.18                      & 70.40                       & 70.62                        & 70.08                        \\ \bottomrule
\end{tabular}
}
\caption{Comparison of momentum values $m$ in pre-training stage on VQA Dev-set.}
\label{tab:momentum}
\end{table}

\begin{table}[!htbp]
\centering
\setlength{\abovecaptionskip}{6pt}
\setlength{\belowcaptionskip}{0pt}
\scalebox{0.85}{
\begin{tabular}{c|ccc}
\toprule
\multirow{2}{*}{Droplayer Policy} & \multirow{2}{*}{\begin{tabular}[c]{@{}c@{}}Keynet w/o Droplayers\\ (Epoch 1-40)\end{tabular}} & \multirow{2}{*}{\begin{tabular}[c]{@{}c@{}}Keynet with Even Droplayers\\ (Epoch 1-40)\end{tabular}} & \multirow{2}{*}{\begin{tabular}[c]{@{}c@{}}Keynet with Delayed Even Droplayers\\ (Epoch 21-40)\end{tabular}} \\
                        &                                                                                               &                                                                                                     &                                                                                                                 \\ \hline
Acc\%                     & 70.27                                                                                         & 70.06                                                                                               & 70.62                                                                                                           \\ 
\bottomrule
\end{tabular}
}
\caption{Comparison of different droplayer policies on VQA Dev-set.}
\label{tab:droplayer}
\end{table}

\begin{table}[!htbp]
\centering
\setlength{\abovecaptionskip}{6pt}
\setlength{\belowcaptionskip}{0pt}
\scalebox{0.9}{
\begin{tabular}{c|ccc}
\toprule
\multirow{2}{*}{Methods}                                             & \multirow{2}{*}{VQA--Dev-set} & \multirow{2}{*}{GQA--Dev-set} & \multirow{2}{*}{NLVR2--Dev-set} \\
                                                                    &                      &                      &                        \\ \hline
No Vision Task                                                      & 66.30                & 57.10                & 50.90                  \\ 
Feature Regression                                                  & 69.10                & 59.45                & 72.89                  \\ 
\begin{tabular}[c]{@{}c@{}}Feature Regression + Label\end{tabular} & 69.90                & 59.80                & 74.51                  \\ 
Contrastive Learning                                                & 70.62                & 59.21                & 76.47                  \\ \bottomrule
\end{tabular}
}
\caption{Comparison of different loss compositions on dev splits of VQA, GQA and NLVR2.}
\label{tab:loss}
\end{table}

\begin{table}[!htbp]
\centering
\setlength{\abovecaptionskip}{6pt}
\setlength{\belowcaptionskip}{0pt}
\scalebox{0.9}{
\begin{tabular}{c|ccc}
\toprule
\multirow{2}{*}{\begin{tabular}[c]{@{}c@{}}Momentum value for fine-tuning\end{tabular}} & \multirow{2}{*}{$m$ = 0.9995} & \multirow{2}{*}{$m$ = 0.99995} & \multirow{2}{*}{$m$ = 0.99997} \\
                                                                                            &                             &                              &                              \\ \hline
\begin{tabular}[c]{@{}c@{}}NLVR2--Dev-set\\ (1 Epoch = 2699 Iterations)\end{tabular}                   & 76.47                       & 72.19                        & -                            \\ 
\begin{tabular}[c]{@{}c@{}}VQA--Dev-set\\ (1 Epoch = 19753 Iterations)\end{tabular}                    & 70.31                      & 70.62                        & -                            \\ 
\begin{tabular}[c]{@{}c@{}}GQA--Dev-set\\ (1 Epoch = 33595 Iterations)\end{tabular}                    & -                           & 58.98                        & 59.21                       \\ 
\bottomrule
\end{tabular}
}
\caption{Comparison of momentum values $m$ in fine-tuning stage on the dev splits of NLVR2, VQA, and GQA.}
\end{table}

\subsection{Ablation Studies and Analyses of CVLP}
\noindent {\bf Effects of Momentum Value.}
Momentum controls the weight movement in the key network. A large momentum will result in slow drift of features. From Table~\ref{tab:momentum}, we can infer that a larger momentum results in a better performance on VQA because the feature drift is reduced. However, as the momentum grows to 1, the performance can drop significantly because the weight in the key network will stop to accept new information. In our experiments, we empirically determine a proper  
value for the momentum $m$ as $m = 1 - 1/I$, where $I$ is the iteration step in one epoch.

\noindent {\bf Effects of Droplayer Policy.}
Due to the powerful discrimination ability, contrastive learning easily overfits the training data. Our droplayer in the key network is an important technique to tackle the over-fitting problem of contrastive learning. As shown in Table~\ref{tab:droplayer}, the key network with droplayer applied on the even layer decreases the performance. By applying a delayed droplayer policy, which takes effects after 20 epochs on even layers, the performance is significantly improved over the key network without droplayer. This experiment demonstrates the effectiveness of our proposed droplayer technique. 

\noindent {\bf Effects of Loss Composition.}
In Table~\ref{tab:loss}, we perform an ablation study on different loss combinations. No vision task performs visual-linguistic pretraining without adding masks on the visual features. By adding feature regression over masked visual regions, we can achieve improved performance. By adding feature regression and label classification, the results can be improved even further. After replacing feature regression and label classification loss with contrastive loss, we achieve improved performance over the three LXMERT variants on VQA and NLVR2. The performance on GQA is worse than LXMERT. This consolidates our claim that contrastive learning can perform better when the gap between pretraining and finetuning is large.

\begin{figure*}[htbp]
 \centering
 \includegraphics[width=0.95\textwidth]{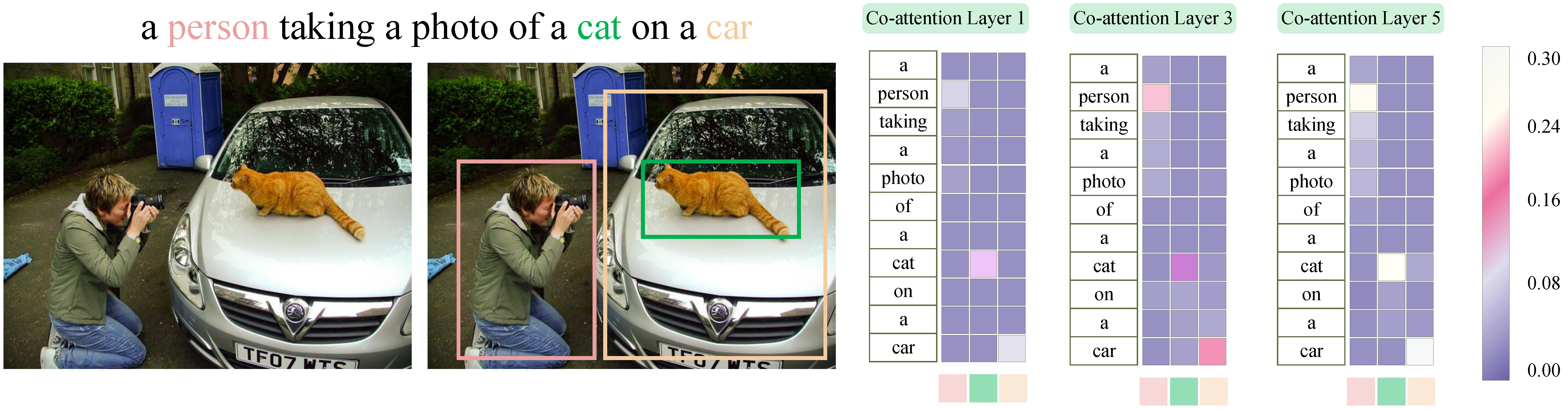}
 \caption{Illustration of attention weights in odd layers of the Co-attention Encoder. The lighter the color, the greater the weight of attention.}
 \label{vision}
\end{figure*}

\paragraph{Visualizing CVLP Encoder.}
In Figure ~\ref{vision}, we visualize the attention weights in odd layers (i.e., the 1st, 3rd, 5th layers) of the Co-attention Encoder in CVLP. We can see that as the layer grows, the attention weight which indicates correct word-bounding box matching also increases gradually.

\section{Conclusion}
In this paper, we propose a contrastive learning based pretraining approach for visual-linguistic representation. CVLP is not biased by the visual features pretrained on Visual Genome and can achieve superior performance, particularly when there is a substantial gap between pretraining and the downstream task. Extensive experiments and ablation studies over VQA, NLVR2, as well as GQA demonstrate the effectiveness of CVLP.

\section*{Broader Impact}
Deep learning algorithms frequently achieve superior performance on supervised tasks. However, due to their large numbers of parameters, they often require high quality and abundant training labels. Such annotation can be time-consuming and expensive. Our proposed CVLP can perform high-quality representation learning based on self-supervised pretext tasks. We believe our research can help many deep learning applications and decrease the overall cost to train and deploy a deep learning system. Large-scale pretraining with models that can cope with a domain gap have the potential to reduce possible energy usage, as one does not need to train a model from scratch for new domains. Moreover, self-supervised learning can allow us to learn from more available unlabeled data, enabling us to mitigate the well-known problems of bias and fairness in human annotations. Still, it remains important to consider the distribution of the unlabeled data to avoid biases in the model.

{\small
\bibliographystyle{IEEEbib}
\bibliography{refs}
}

\end{document}